\documentclass[%
 aip,
 amsmath,amssymb,
 reprint,%
]{revtex4-1}

\usepackage{graphicx}
\usepackage{dcolumn}
\usepackage{bm}

\usepackage[utf8]{inputenc}
\usepackage[T1]{fontenc}
\usepackage{mathptmx}
\usepackage{etoolbox}
\usepackage[backref]{hyperref} 
\usepackage{color}
\usepackage{subfigure}
\usepackage{graphicx}

\makeatletter
\def\@email#1#2{%
 \endgroup
 \patchcmd{\titleblock@produce}
  {\frontmatter@RRAPformat}
  {\frontmatter@RRAPformat{\produce@RRAP{*#1\href{mailto:#2}{#2}}}\frontmatter@RRAPformat}
  {}{}
}%
\makeatother
\begin{document}

\preprint{AIP/123-QED}

\title{A Note on Learning Rare Events in Molecular Dynamics using  \\LSTM and Transformer}
\author{Wenqi Zeng}

\affiliation{Department of Mathematics, Hong Kong University of Science and Technology}
\author{Siqin Cao}
\affiliation{Department of Chemistry, Hong Kong University of Science and Technology}

\author{Xuhui Huang}
\email[Authors to whom correspondence should be addressed: ]{xuhuihuang@ust.hk or yuany@ust.hk}
\affiliation{Department of Chemistry, Hong Kong University of Science and Technology}
\affiliation{Department of Chemical and Biological Engineering, Hong Kong University of Science and Technology}

\author{Yuan Yao}
\affiliation{Department of Mathematics, Hong Kong University of Science and Technology}
\affiliation{Department of Chemical and Biological Engineering, Hong Kong University of Science and Technology}

\date{\today}

\begin{abstract}
Recurrent neural networks for language models like long short-term memory (LSTM) have been utilized as a tool for modeling and predicting long term dynamics of complex stochastic molecular systems. Recently successful examples on learning slow dynamics by LSTM are given with simulation data of low dimensional reaction coordinate. However, in this report we show that the following three key factors significantly affect the performance of language model learning, namely dimensionality of reaction coordinates, temporal resolution and state partition. When applying recurrent neural networks to molecular dynamics simulation trajectories of high dimensionality, we find that rare events corresponding to the slow dynamics might be obscured by other faster dynamics of the system, and cannot be efficiently learned. Under such conditions, we find that coarse graining the conformational space into metastable states and removing recrossing events when estimating transition probabilities between states could greatly help improve the accuracy of slow dynamics learning in molecular dynamics. Moreover, we also explore other models like Transformer, which do not show superior performance than LSTM in overcoming these issues. Therefore, to learn rare events of slow molecular dynamics by LSTM and Transformer, it is critical to choose proper temporal resolution (i.e., saving intervals of MD simulation trajectories) and state partition in high resolution data, since deep neural network models might not automatically disentangle slow dynamics from fast dynamics when both are present in data influencing each other.
\footnote{The code is available at: \url{https://github.com/Wendysigh/LSTM-Transformer-for-MD}}
\end{abstract}

\maketitle

\section{\label{Introduction}Introduction}

Molecular dynamics (MD) has been widely used as a tool to study conformation dynamics in recent years. In a molecular dynamics (MD) simulation, molecules are modelled with interacting beads and chemical bonds, where the interactions between beads are modelled with force field. The result of MD simulations are normally stored as snapshots of atomic coordinates of the simulated molecular system on a discrete time sequence,  which are called trajectories. The original dynamics of MD systems evolved in a high dimensional space that contains a huge degree of freedom. However, for most cases only the slow dynamics are of interest as they often underline the function of biological macromolecules. This provides possibility to model the MD dynamics with low dimensions, e.g., reaction coordinates or a state model like Markov State Model (MSM) \cite{chodera2014markov,husic2018markov,prinz2011markov,malmstrom2014application,bowman2013introduction,chodera2007automatic,pan2008building,sarich2010approximation,noe2013variational,wu2020variational,weng2020revealing,zeng2018harnessing,zhang2016simulating,bowman2010network,yao2013hierarchical,da2013two,unarta2021role,morcos2010modeling,huang2009rapid,buchete2008coarse,noe2009constructing,bowman2011taming,buch2011complete,silva2011role,noe2007hierarchical,bowman2010enhanced}. In these models, the dynamics of  trajectories are stored in very low dimensions (e.g. in a state model the dynamics are stored as a sequence of state index), and thus provides possibility to model the dynamics with other non-simulation methods due to their sequential characteristics. These models are required to make correct predictions of the future state of the molecular systems based on information of the past with appropriate length, e.g. a state model of molecular dynamics system will produce a series of numbers with sequence over time. On the other hand, natural human language is composed of sequential states that conform to a certain logic or rule, which may also be similar to predict molecular dynamics. In recent years, deep learning recurrent neural network methods such as gated recurrent unit (GRU) \cite{cho2014learning}, long short-term memory (LSTM) \cite{hochreiter1997long} and their variants have shown great potential in processing sequentiality \cite{lukovsevivcius2009reservoir}, and there are now studies on using them to analyse trajectories from simulation systems \cite{eslamibidgoli2019recurrent}  \cite{pathak2018model}.

Back to the application of recurrent neural network to molecular dynamics, related researches are still limited. A conservative approach is to incorporate LSTM into the numerical integrator that solves Newton’s equations in molecular dynamics simulations \cite{kadupitiya2020deep}. Another applies the recurrent neural network directly onto the low dimensional trajectories and predicts the next token in the sequential data \cite{tsai2020learning}. They proved the training under cross-entropy loss is equivalent to learning a path entropy and captured both Boltzmann statistics and kinetics. In this work, the authors project their MD simulation trajectories onto a one-dimension reaction coordinate and further discretized the MD conformations by equal distance binning.  Pre-processing of MD simulation trajectories to low dimension has been shown to render the LSTM model effective to learn the rare events.  However, the applicability of the LSTM and other language models  directly on high-dimensional data haven't been extensively examined.

In this work, we examined the performance of the LSTM and Transformer model in two scenarios on the alanine dipeptide system: (a). Pre-processing of the MD simulation trajectories by projecting them onto 2 torsional angles: $\phi$ and $\psi$.  (b). Directly decompose MD simulation trajectories into states using the root mean square displacement (RMSD) distances without any pre-processing of the high-dimensional MD data. The performance of LSTM on slow dynamics prediction can severely drop if the saving interval of the MD simulation trajectories is small with both ${\rm alanine}_{\phi \psi}$ and high resolution data ${\rm alanine}_{\rm RMSD}$. To address such an issue, we investigate some methods to capture the rare events in long sequence data. Effective approaches include kinetic lumping \cite{li1989general} in physical chemistry to do different partition on states and remove recrossing \cite{wang2009recrossing} to reduce fast dynamics noise in the input data and enhance slow dynamics signal. On the other hand, Transformer models are not effective in learning slow dynamics as rare events in long sequences, due to that both fast and slow dynamics in training data entangled deep language models in efficiently capturing the latter.

\section{Methods} \label{sec:method}
The basic principle of the task is that the trajectory of molecular dynamics can be discretized into sequential data, where the prediction of the current state is related to the known states of the past. This is similar to natural language processing like LSTM or Transformer. To further introduce these deep learning models, we roughly divide the whole workflow into three parts. Firstly, an embedding layer is used to represent raw input trajectory as vectors \textbf{X} by multiplying learnable weights, and pass \textbf{X} through next part to generate high dimensional representation \textbf{h}. Finally, a fully connected layer will map the learned representation \textbf{h} to the predicted probability for the current state. Among the three sequential part, the main difference between LSTM and Transformer lies in the second part, learning representation \textbf{h}.

\begin{figure}[htbp]
\centering
\subfigure[LSTM]{
\includegraphics[width=5.5cm]{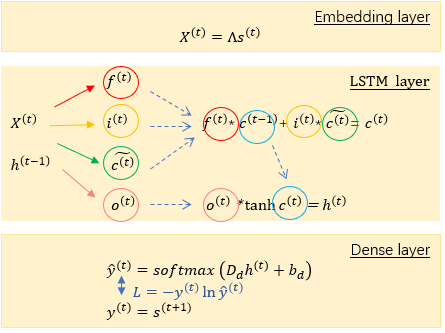}
}
\quad
\subfigure[Transformer with n encoders and n decoders]{
\includegraphics[width=5.5cm]{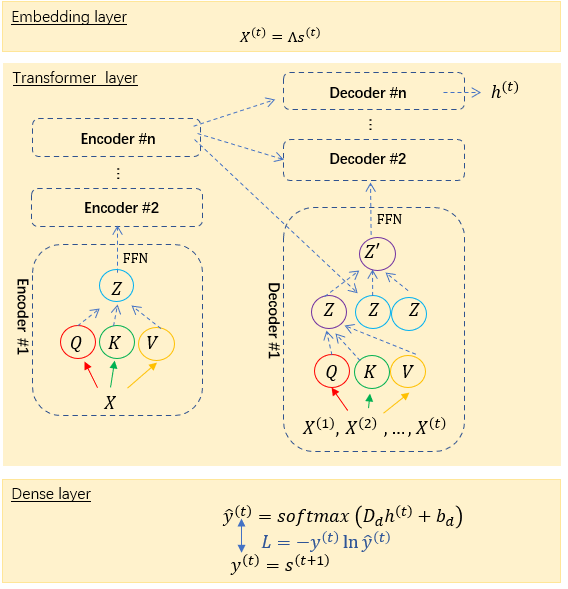}
}
\caption{Procedure for LSTM and Transformer}
\label{model}
\end{figure}

As in Figure~ \ref{model} (a), denote $\boldsymbol{X}_t$ as the $t_{th}$ state input for the LSTM layer. Each $\boldsymbol{X}_t$ generates $\boldsymbol{h}_t$ from LSTM layer. The LSTM itself consists of the following elements: the forget gate $\boldsymbol{f}_t$, the input gate $\boldsymbol{i}_t$, the output gate $\boldsymbol{o}_t$, the cell state $\boldsymbol{c}_t$, and $\boldsymbol{h}_t$ which is the hidden state vector and output from the LSTM layer. Each gate processes information in different aspects. Briefly, the forget gate decides which information to be erased, the input gate decides which information to be written, and the output gate decides which information to be read from the cell state to the hidden state. The relations among these elements can be written as follows:

$$\boldsymbol{f}_t=\sigma(\boldsymbol{W}_f \boldsymbol{X}_t+\boldsymbol{U}_f \boldsymbol{h}_{t-1}+\boldsymbol{b}_f)$$
$$\boldsymbol{i}_t=\sigma(\boldsymbol{W}_i \boldsymbol{X}_t+\boldsymbol{U}_i \boldsymbol{h}_{t-1}+\boldsymbol{b}_i)$$
$$\boldsymbol{o}_t=\sigma(\boldsymbol{W}_o \boldsymbol{X}_t+\boldsymbol{U}_o \boldsymbol{h}_{t-1}+\boldsymbol{b}_o)$$
$$\boldsymbol{\tilde{c}}_t=\tanh (\boldsymbol{W}_c \boldsymbol{X}_t+\boldsymbol{U}_c \boldsymbol{h}_{t-1}+\boldsymbol{b}_c)$$
$$\boldsymbol{c}_t=\boldsymbol{f}_t \circ \boldsymbol{c}_{t-1}+\boldsymbol{i}_t \circ \boldsymbol{\tilde{c}}_t$$
$$\boldsymbol{h}_t=\boldsymbol{o}_t \circ \tanh \boldsymbol{c}_t$$
where $\boldsymbol{W}$ and $\boldsymbol{b}$ are the corresponding weight matrices and bias. The operator $\circ$ stands for the Hadamard product.

In LSTM, the calculation of representation of $t_{th}$ state $\boldsymbol{h}_t$ depends on $\boldsymbol{h}_{t-1}$ and $\boldsymbol{c}_{t-1}$ while Transformer use attention mechanism to avoid such dependency. Transformer is consist of encoders and decoders. In an encoder of Transformer of Figure~\ref{model} (b), $\boldsymbol{X}$ will go through a module called “Multi-Head Attention” to get a weighted feature vector $\boldsymbol{Z}$:
$$\boldsymbol{Q}=\boldsymbol{W}_Q \times \boldsymbol{X}$$
$$\boldsymbol{K}=\boldsymbol{W}_K \times \boldsymbol{X}$$
$$\boldsymbol{V}=\boldsymbol{W}_V \times \boldsymbol{X}$$
$$\boldsymbol{Z}=\texttt{softmax} (\frac{\boldsymbol{Q}\boldsymbol{K}^T}{\sqrt{d_k}})\boldsymbol{V}$$
where $\boldsymbol{W}$ are the corresponding weight matrices and $d_k$ is the dimension of $\boldsymbol{K}$ vectors.

After obtaining $\boldsymbol{Z}$, it will be sent to a Feed Forward neural network with firstly an activation function as ReLU and then a linear function with wight matrices $\boldsymbol{W}_1$, $\boldsymbol{W}_2$ and bias $\boldsymbol{b}_1$, $\boldsymbol{b}_2$.

$$\texttt{FFN} (\boldsymbol{Z}) = \max (0,\boldsymbol{Z} \boldsymbol{W}_1+\boldsymbol{b}_1)\boldsymbol{W}_2+\boldsymbol{b}_2$$

The $\boldsymbol{K}$ and $\boldsymbol{V}$ vectors from encoder will be combined with the $\boldsymbol{Q}$ vectors in decoder to output $\boldsymbol{Z}$ and a final representation $\boldsymbol{h}=\texttt{FFN} (\boldsymbol{Z})$ to be mapped into categorical probability by the linear layer.

LSTM can be very effective for data with sequential characteristics with the ability to mine timing and semantic information in domains including but not limited to speech recognition, machine translation, and timing analysis. There is no doubt these variants can learn long-term information \cite{karpathy2015unreasonable}, but they also suffer from the memory loss despite of the existence of gating mechanism \cite{zhao2020rnn}. Essentially, vanilla LSTM only expects to influence future decisions with past states, which is known as “unidirectional”. In the training process, we could adopt a ``bidirectional'' trick that both the past and future states are considered, which can provide more sufficient contents to encode. The third issue comes from unparallelizable recurrence in these models, which limits the acceleration of parallel computing by GPU. Transformer \cite{vaswani2017attention} solves the above three limitations through attention mechanism, that is, by calculating the similarity scores between each word and other words, the distance between any two words is 1, which is neither restricted by long-distance dependence nor subject to unidirectional operations like LSTM. Based on the attention mechanism, Transformer is more suitable for GPU acceleration without using sequence aligned LSTM. 

\section{Results} \label{sec:experiment}
\subsection{Datasets, Settings and Evaluation Metrics}
Here we choose conformational dynamics of alanine dipeptide in water to test LSTM and Transformer. The full data set of alanine dipeptide includes 100 trajectories of an alanine-dipeptide and 888 water molecules, the length of each trajectories is 10ns and the snapshots are saved every 0.1ps. In the original data set, alanine dipeptide contains 22 atoms and 66 cartisian coordinates. As most degrees of freedom are related with fast dynamics (like vibration of chemical bonds), we derived the original trajectories into two low-dimensional data sets: the ${\rm alanine}_{\phi \psi}$ data set and ${\rm alanine}_{\rm RMSD}$ data set. In the ${\rm alanine}_{\phi \psi}$ data set, we used the torsional angles $\phi$ and $\psi$ to capture the conformation changes of backbone. In the ${\rm alanine}_{\rm RMSD}$ data set, the conformation space is split into 100 states using $k$-center clustering method \cite{zhao2013fast} that finds an approximate $\epsilon$-cover of samples \cite{Sun2008,Yao_JCP09,Yao_JCP13} according to their RMSD distance of heavy (i.e. non-hydrogen) atoms. In the first kind of models, the  $\phi$-$\psi$ based model, each frame of the input data set only includes the $\phi$ or $\psi$ angle; while in the RMSD based model, each frame of the input data set only includes a state index ranging from 0 to 99. Here our $\phi$-$\psi$ based model is consistent with the model in a previous paper \cite{tsai2020learning}, where we also adopted the 20-bin states corresponding to the values of $\phi$ or $\psi$.

For LSTM, we set the embedding dimension to 128 and LSTM units to 1024. The trajectories were batched into sequences with length of 100 and batch size of 64. For Transformer, we change the embedding size to 512 and output dimension to 2048 with 8 heads, 2 stacks, 0.1 drop-out ratio. Specifically for Transformer, we apply Adam optimizer with $\beta_1=0.9$, $\beta_2=0.98$ and $\epsilon=10^{-9}$. We varied the learning rate over the course of training according to the formula: $lrate=d_{model}^{-0.5} \cdot \min(\text{step-num}^{-0.5},\ \text{step-num} \cdot \text{warmup-steps}^{-1.5})$. This corresponds to increase the learning rate linearly for the first warmup-up training, and decrease it thereafter proportionally to the inverse square root of the step number. Here we use warmup-steps = 4000. 

In generating process we use the trained model to recursively generate future predictions by appending the predicted values to the original sequence and shift the old values. We generate 100 trajectories with sequence length of 10,000 and do 50 times bootstrap to evaluate.   

As for evaluation, we consider free energy ($-\log (\text{population})$) and other evaluation on kinetics like Implied Time Scales (ITS) which is calculated from the Markov model eigenvalues, and Mean First-Passage Time (MFPT). In MSM, the ITS is used to estimate the Markovian lag time, which is also representative to the order of magnitude of dynamics:
\begin{equation}
	{\rm ITS}_i = - \frac{\tau}{\ln\lambda(\tau)_i} 
\end{equation}
where $\tau$ is the lag time to compute auto-correlation function of states $C(\tau)=\overline{x(t)x(t+\tau)}$, $x(t)$ is the state index at time $t$, and $\lambda_i(\tau)$ is the $i$-th eigenvalue of the transition probability matrix $T(\tau)$. The transition probability matrix is a normalized auto-correlation function: $T_{ij}(\tau)=C_{ij}(\tau)/\sum_iC_{ij}(\tau)$.

The MFPT denotes the transition time between each pair of states. For a Markov chain, the MFPT can be calculated by \cite{sheskin1995computing}:
\begin{equation}
	t_{ij} = \tau + \sum_kT_{ik}t_{kj}
\end{equation}
where the transition probability matrix $T$ needs to be Markovian.
\subsection{Experiments on LSTM and Transformer}
In this section we used the conformational dynamics of alanine dipeptide in water to test deep learning models (i.e. LSTM and Transformer). Both these two data sets contain the degree of freedom of slow dynamics, and deep learning models are used to reconstruct the dynamics in the reduced dimensionality. In the following part of this section, firstly LSTM is tested on both data sets. Then the  Transformer algorithm is presented as a possible alternative to the LSTM model.

\subsubsection{Results on ${\rm alanine}_{\phi \psi}$}
In this part we show that with 1ps or 2ps saving intervals, one can learn LSTM well for the 20-state on $\phi-\psi$ plane in terms of both thermodynamic equilibrium distribution and the long term dynamic behavior. However, when the saving interval varies to 0.1ps, LSTM fails to capture the long term behaviour.

\begin{figure}[htbp] 
\centering
\includegraphics[scale=0.4]{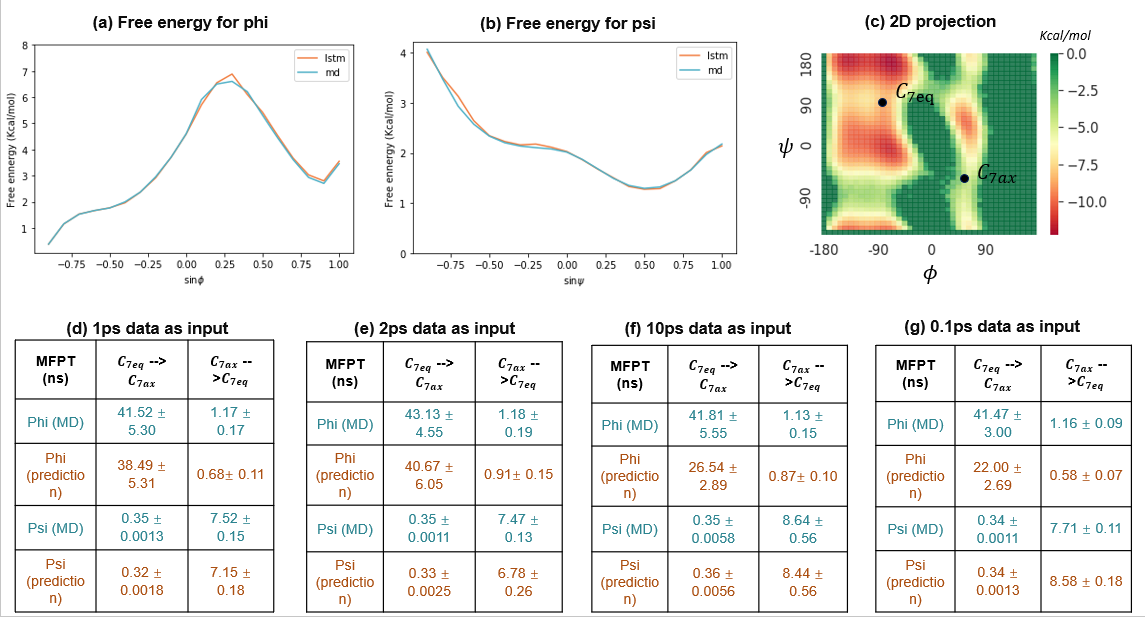}
\caption{(a) and (b) are free energy for $\phi$ and $\psi$ on 0.1ps saving interval respectively. (c) 2D projection onto $\phi$-$\psi$ angles (d) MD simulation data with 1ps saving interval (e) MD simulation data with 2ps saving interval (f) MD simulation data with 10ps saving interval (g) MD simulation data with 0.1ps saving interval. (d)(e) differs obviously from (f)(g)}
\label{mfpt lstm}
\end{figure}

The LSTM can not only correctly predict the thermodynamic properties in Figure~\ref{mfpt lstm} (a)(b), but also be able to correctly capture the slow dynamics of the ${\rm alanine}_{\phi \psi}$ data sets.  As can be seen from the results in Figure~\ref{mfpt lstm} (d)(e), the dynamic information can be captured by LSTM and then mimicked in predicted trajectories at saving interval 1ps or 2ps, on both $\phi$ and $\psi$ angle. Therefore, it is feasible to learn and model slow dynamics for alanine dipeptide by deep learning language models like LSTM. The only requirement for using LSTM is sufficient amount of data, which is already available and circumvents highly resource consumption in prediction by molecular dynamic simulation system. In a similar way, LSTM may not work fine on the data sets with 10ps saving interval, as each of our trajectories has 10ns data and the 10ps saving interval data sets may suffer from insufficiency of data.

However, if we use shorter saving interval like 0.1 ps and more frames would be needed to capture the slow dynamics, LSTM would not produce accurate results comparing to the original MD simulations (see Figure~\ref{mfpt lstm} (g) ). This indicates that the saving interval will indeed have an adverse impact, especially in moving from metastable state $C_{7eq}$ to $C_{7ax}$ with respect to angle defined in Figure~\ref{mfpt lstm} (c)$\phi$.

\subsubsection{Results on ${\rm alanine}_{\rm RMSD}$}
Reaction coordinates, which distinguish ${\rm alanine}_{\rm RMSD}$ and ${\rm alanine}_{\phi \psi}$, and saving interval serve as two influencing factors in this part. When the data set comes to ${\rm alanine}_{\rm RMSD}$ , with 1ps saving interval, the performance of LSTM is acceptable although it is not as good as on ${\rm alanine}_{\phi \psi}$. Still, LSTM performs much worse on 0.1ps saving interval than ${\rm alanine}_{\phi \psi}$, failing to capture the slowest motion. 

\begin{figure}[htbp] 
\centering
\includegraphics[scale=0.4]{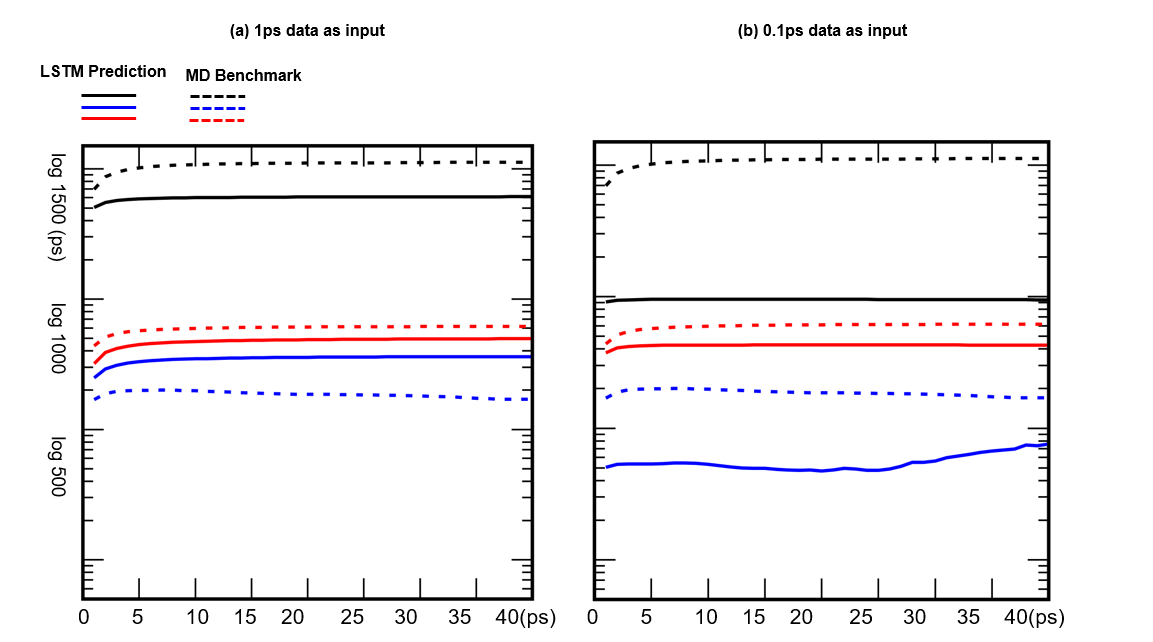}
\caption{ITS on $alanine_{RMSD}$ for LSTM with sequence length=100. (a) result on 0.1ps saving interval. (b) result on 1ps saving interval. 1ps data outperforms 0.1ps data.}
\label{its rmsd}
\end{figure}

On $alanine_{RMSD}$ dataset, the slowest motion, that is, the first ITS represented by the black line, is the dominant one and can be captured with saving interval as 1ps. In Figure~\ref{its rmsd} (a), the first ITS of predicted 1ps saving interval dataset is around 700ps while that of MD trajectories is around 1000ps. LSTM is able to be applied on a more complicated data set as ${\rm alanine}_{\rm RMSD}$ considering ${\rm alanine}_{\rm RMSD}$ contains all conformational dynamics of alanine dynamics, which is more challenging than the one-dimensional data sets ${\rm alanine}_{\phi \psi}$. However, the negative impact of small saving interval like 0.1ps is still inevitable. The gap between LSTM prediction and benchmark gets enlarged on 0.1ps saving interval dataset as shown in Figure \ref{its rmsd} (b). 

\subsubsection{Improvement on 0.1ps saving interval}
Based on results of previous two datasets, LSTM works at saving interval 1ps but fails when the saving interval is 0.1ps with high temporal resolution, and its performance also gets affected by the pre-processing performed on MD simulation trajectories. We proposed several effective solutions based on the idea to suppress the fast dynamic modes as noises in favor of capturing slow motions as rare events. 

As we have seen, LSTM performs worse when predicting slow dynamics as rare events on data sets with 0.1ps saving interval than the low resolution of 1ps. Therefore, rare events may not be well handled in the high resolution dataset for LSTM. To explore possible reasons, we further study the difference between 0.1ps and 1ps saving interval on the ${\rm alanine}_{\rm RMSD}$ data set. There are quick transitions between states in 0.1ps data whose frequency of occurrence is low, which makes LSTM hard to tell the signals from noise. As a result, LSTM tends to learn some high frequency event or noise, and ignore the low frequency event.

Considering that the rare events as slow dynamics are very likely to be masked in the 0.1ps data, we choose the following ways to tackle. 

\textbf{Use kinetic lumping methods to find metastable states}. In $k$-center clustering, which is sensitive to noise, states are not metastable and there exists fast intra-metastable disturbance which adds noise to the trajectories to confuse the language models. Lumping such states toward metastable macro-states thus reduce fast dynamics and enlarge the signal, the slow dynamics. For example, by turning the 100 state model into a 4 metastate model, the fast intra-metastable dynamics will be removed from the data set, and  most rare events can be included in new 4-states transitions. Many lumping methods can be used here, e.g. the Hierarchical Nystr\"{o}m method \cite{Yao_JCP13}, Perron-Cluster Cluster Analysis (PCCA) method \cite{deuflhard2005robust}, PCCA+ \cite{roblitz2013fuzzy}, Maximum PowerPoint (MPP) \cite{esram2007comparison}. Here we adopted PCCA+ to do lumping.

\textbf{Remove recrossing}. The recrossing refers to some high frequently jumps back and forth, which can be seen as noise and removed from the data set before processed with LSTM. 

The results in Table~\ref{solution} (a) shows lumping method and recrossing removal has positive effect on improving prediction by LSTM while changes on input data expression does not see improvement. Therefore, denoising by reducing the fast dynamics helps language models to learn the slow dynamics appeared as rare events.

Finally we note that another way of using a different expression of input data does not work well in our experiments. In this different expression, the original input data is condensed into data points in the form of “states-consecutive length”, where the former “states” stands for the interstate transition and latter “consecutive length” for intrastate transition. For example, original data like "4,4,4,4,4,3,3,3,2,2,2,2,..." will be transformed into "4-5, 3-3,2-4,..." in a short format of ``state-length". But this does not help to reduce the noise shown in Table~\ref{solution} (a). 

\subsubsection{Ablation Studies}
In this part we present two ablation studies on sequence length and Transformer, both of which, however, are not effective in improving the 0.1ps case.

\textbf{Variations on sequence length}.
Despite that we have more detailed information on the 0.1ps data, it performs worse than other larger saving interval. There are two speculations about this counter intuitive phenomenon: one is each long sequence in batched training data has introduced large amount of fast dynamics, it may be helpful to shorten the sequence length to 15, 20, 50 to reduce the fast dynamics in processing each sequence. The other is on the contrary. With the same sequence length, compared with 1ps,  events are even sparser in 0.1ps data because slow dynamics requires more frames to capture. For the second conjecture, we increase the sequence length to 1000. Nevertheless, this approach does not see any effect because the sequence length are indeed not restricted under the setting of ``stateful" LSTM as shown next.

In comparison, we also bring in a parameter in LSTM called ``stateful" versus ``stateless". ``Stateful" indicates LSTM will remember the content of the previous batches and ``stateless" will throw them away. Stateful LSTM will find the dependencies among all input sequences and may have longer memory than input sequence length. From results in Table~\ref{solution} (b) the sequence length is similar to a hyperparameter like learning rate or batch size that has an impact on the model, but it does not greatly improve the performance on 0.1ps $alanine_{RMSD}$. Moreover, stateful LSTM outperforms stateless LSTM on this task and we keep stateful as basic setting in all LSTM experiments.

\begin{table}[hbp]    
\centering
\includegraphics[scale=0.6]{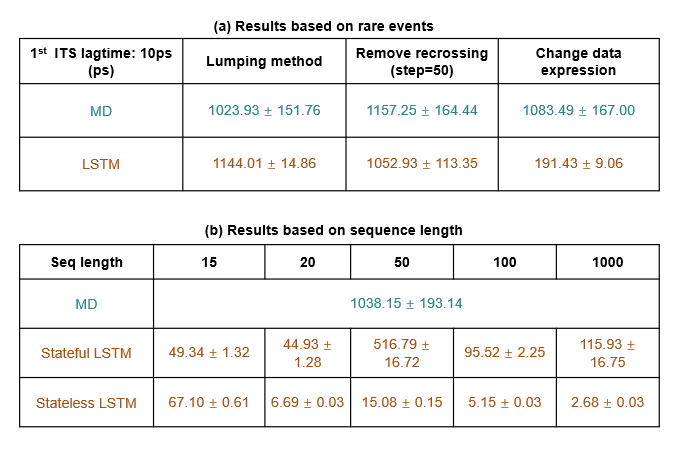}
\caption{(a) 1st ITS on 0.1ps $alanine_{RMSD}$  for solution based on rare events.  (b)1st ITS on 0.1ps $alanine_{RMSD}$  for solution based on sequence length. Different sequence length does not help to improve. Stateful predictions outperforms stateless predictions. Using lumping method and removing recrossing see some improvement.}
\label{solution}
\end{table}

\begin{figure}[htbp] 
\centering
\includegraphics[scale=0.4]{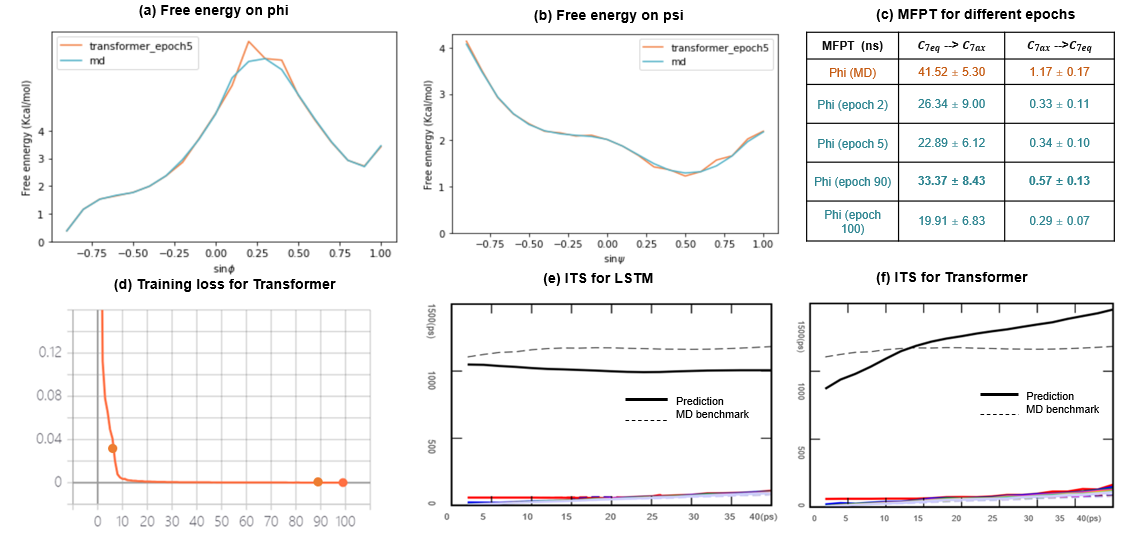}
\caption{(a) (b) are free energy on phi and psi by transformer respectively. (c) MFPT for predicted trajectories by Transformer different epochs. (d) corresponding training loss for epochs used in (c). (e) ITS for LSTM (f) ITS for Transformer on $alanine_{RMSD}$ dataset with saving interval=1ps. In total free energy landscape is barely satisfactory. ITS of LSTM predictions will converge while Transformer's not.}
\label{its2}
\end{figure}

\textbf{Experiments with Transformer}.
This idea originates from some intrinsic drawbacks of LSTM model like memory loss in dealing with long sequence. Here long sequence issue refers to the fact that the dynamic can stay in one state for a long time, where LSTM could suffer from memory leak despite of the existence of gating mechanism, thus we apply Transformer to explore whether it outperforms LSTM on long sequence matter. 
On ${\rm alanine}_{\phi \psi}$, free energy landscape is barely satisfactory in Figure~\ref{its2} (a)(b) but the kinetic information of predictions by Transformer like ITS or MFPT is largely affected by saving interval and model from different epochs although the training loss has already converged to a small number, as shown in Figure~\ref{its2} (c)(d). The performance on Transformer fluctuate so much that most of them deviate from the groundtruth. On ${\rm alanine}_{\phi \psi}$, once the saving interval is increased to, for example, 30 times to 3ps, the trajectories predicted by Transformer will show completely different properties from MD trajectories. Also, on 1ps saving interval ${\rm alanine}_{RMSD}$ shown in Figure~\ref{its2} (e)(f), the ITS of Transformer predictions keeps growing while that of LSTM predictions will converge close to the benchmark. On the whole, the performance of Transformer does not reflect the superiority of attention mechanism over LSTM especially with large saving interval.

Next, we also tested the impact of unidirectional and bidirectional encoding on Transformer. Due to the two-way encoding used by original Transformer encoder, we applied a triangular mask to cover subsequent positions to achieve the purpose of unidirectional encoding. However, since the original bidirectional Transformer performs too bad to be referenced, although the unidirectional results are quite different from bidirectional ones still it is not rigorous to conclude which is better in Table~\ref{uni bi}.

\begin{table}[htbp] 
\centering
\includegraphics[scale=0.6]{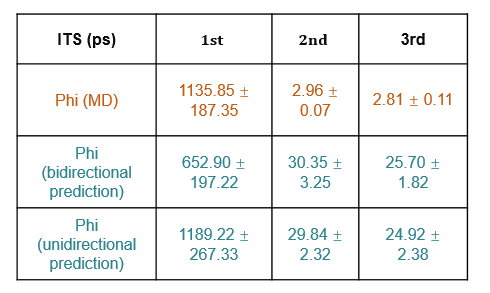}
\caption{(a)(b): MFPT on $\phi$ and $\psi$ respectively (c)(d): ITS on $\phi$ and $\psi$ respectively for unidirectional and bidirectional Transformer predictions on $alanine_{\phi \psi}$ with saving interval=0.1ps. Different but both poor performance.}
\label{uni bi}
\end{table}

\section{Conclusions}
Learning rare events as slow dynamics prediction by neural language models are attractive due to the success of deep learning in recent years. Yet, in this report we show that there are three key factors for the performance of deep learning methods like LSTM and Transformer on learning the rare events in MD simulation trajectories: namely reaction coordinate, temporal resolution like saving interval and state partition.

Reaction coordinate differentiate our two datasets, ${\rm alanine}_{\phi \psi}$ and ${\rm alanine}_{\rm RMSD}$. With one-dimensional sequential data on angle $\phi$ or $\psi$, ${\rm alanine}_{\phi \psi}$ is an easier dataset than ${\rm alanine}_{\rm RMSD}$ which contains more complicated transitions. Specifically, LSTM is able to correctly predict the free energy landscape and dynamics for both the ${\rm alanine}_{\phi \psi}$ and ${\rm alanine}_{\rm RMSD}$ data sets when the saving intervals are 1ps or 2ps. However, LSTM works poorly when the save intervals are 0.1ps or 10ps. For a large saving interval like 10ps, the deep learning model may suffer from insufficiency of data. While for a small saving interval like 0.1 ps, the LSTM may suffer from rare events mixing up with fast dynamics and thus cannot capture slow dynamics efficiently.

Although the free energy landscape predicted with 0.1ps saving interval still overlaps with MD, the dynamics (both ITS and MFPT) of LSTM is much faster than the MD simulations. With the intention of calculating long-term dynamic properties from finer transitions like 0.1ps, we applied methods from the perspective of alleviating the failure to predict the slowest dynamics of the system and took effect. Increasing saving intervals and removing recrossing would largely reduce the short-term dynamics that can be regarded as noise, while applying state partition to ${\rm alanine}_{\rm RMSD}$ could also help in view of turning original $k$-center clustering states into metastable ones and reducing fast intrastate transitions. But increasing the complexity of data sets (e.g. change data form) would lead to even worse performance of LSTM.
Then considering the limitation of LSTM on dealing with long sequence data, we brought about Transformer. The experimental results show Transformer cannot surpass LSTM on learning slow dynamics from MD simulation tajectories. The advantages of the attention mechanism, such as the processing of long sequences without memory loss and bidirectional encoding, have not come into force. The possible reason is that stateful LSTM sees information beyond sequence length while Transformer is limited to given sequence length of 100. 

Regarding the effect of learning rare events on slow dynamics prediction using LSTM and Transformer, choosing proper temporal resolution like saving intervals and state partition are critical for deep learning models. In a summary, deep language models such as LSTM and Transformer, are not able to disentangle the slow dynamics from fast dynamics when both are present in the training data, the former playing a role as signal while the latter as noise in learning rare events as slow dynamics. Therefore, a denoising mechanism is helpful to improve the accuracy of rare event learning by language models with complex molecular dynamics data. Under such assistance, LSTM and Transformer could learn slow dynamics more effectively with high dimensional MD simulation data. Moreover, it is desired to develop advanced deep learning models toward multiscale analysis of molecular dynamics data, that can automatically learn and separate time scales in trajectories. Such a great potential of deep learning models in molecular dynamics appliations are our future directions.

\section*{Reference}
\bibliographystyle{unsrtnat}
\bibliography{ref.bib}

\newpage

\section{Supplementary Information}\label{Supplementary Information}
This part contains supplements to more evaluation metrics on experiments in previous text. 

Table ~\ref{its lstm} shows the first ITS of $alanine_{\phi \psi}$ on different saving interval, serving as a supplement to the MFPT results using LSTM. Similar to MFPT, with larger saving intervals, LSTM performs better.

Table~ \ref{stateful_all} shows the effect of sequence length on LSTM by former three ITS. In main text we only showed results on the first ITS. On all three ITS, sequence length does not have dominant effect. 

\begin{figure}[hbp] 
\centering
\includegraphics[scale=0.5]{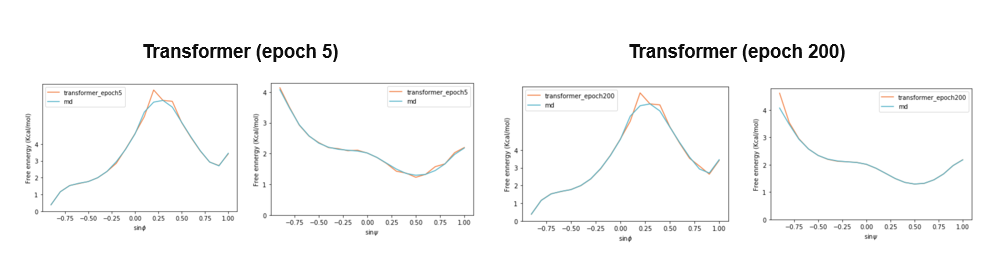}\label{fig.distribution}
\caption{Free energy landscapes for different epochs of Transformer on $alanine_{\phi \psi}$, 1ps. Basically satisfying.}
\label{fe}
\end{figure}

\begin{figure}[hbp] 
\centering
\includegraphics[scale=0.4]{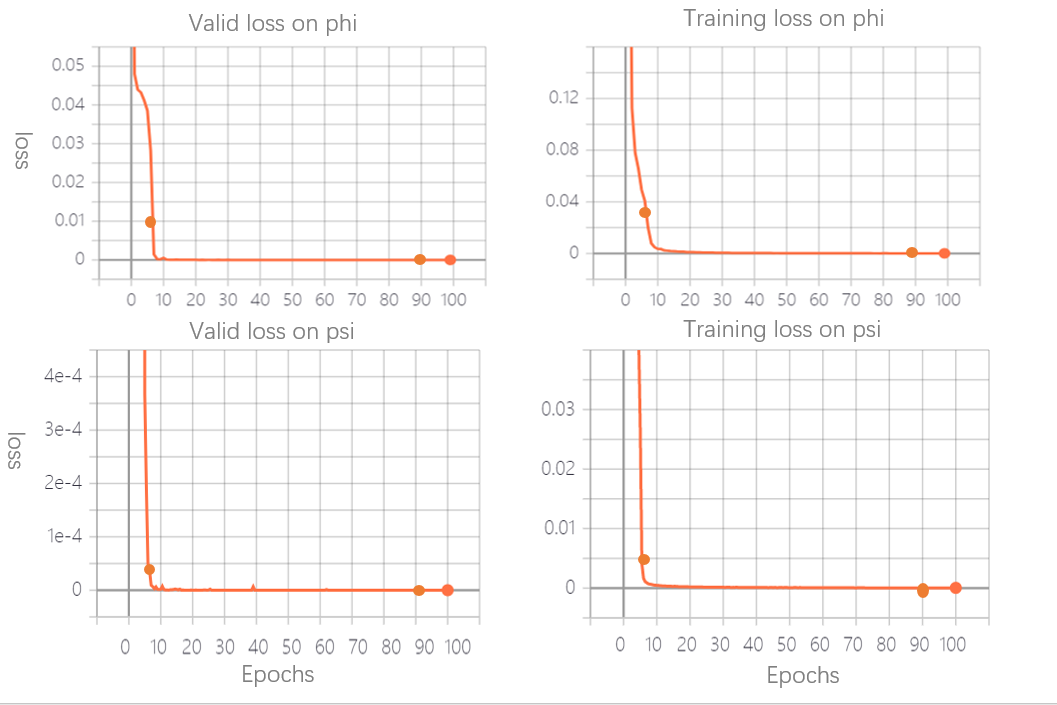}
\caption{Loss plot for Transformer on $\phi$ and $\psi$ respectively.}
\label{loss}
\end{figure}

Figure~\ref{fe} shows different epochs of Transformer model possess robust performance on thermodynamic equilibrium distribution. Table~\ref{mfpt} and Table~\ref{its} contain detailed results  using different epochs of Transformer model or different saving intervals to convey that the long term dynamic behavior is harder to be captured than equilibrium distribution. Figure~\ref{loss} depicts training and validation loss for Transformer to put away worries such as non convergence.

\begin{table}[htbp]  
\centering
\includegraphics[scale=0.6]{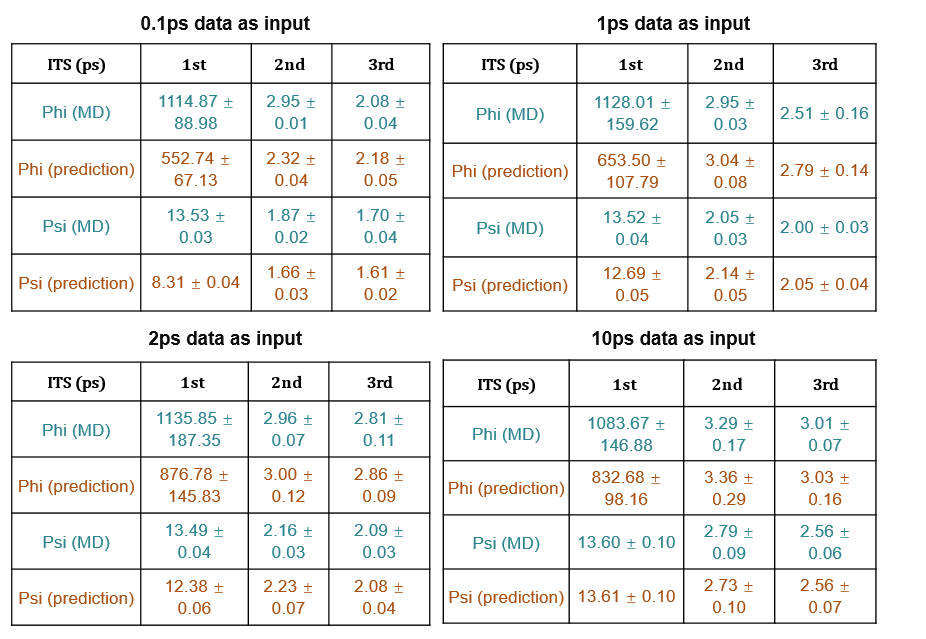}
\caption{ITS on $alanine_{\phi \psi}$ for LSTM with different saving interval. (a) saving interval=0.1ps (b) saving interval=1ps (c) saving interval=2ps (d) saving interval=10ps. Larger saving interval, better performance.}
\label{its lstm}
\end{table}

\begin{table}[htbp]    
\centering
\includegraphics[scale=0.6]{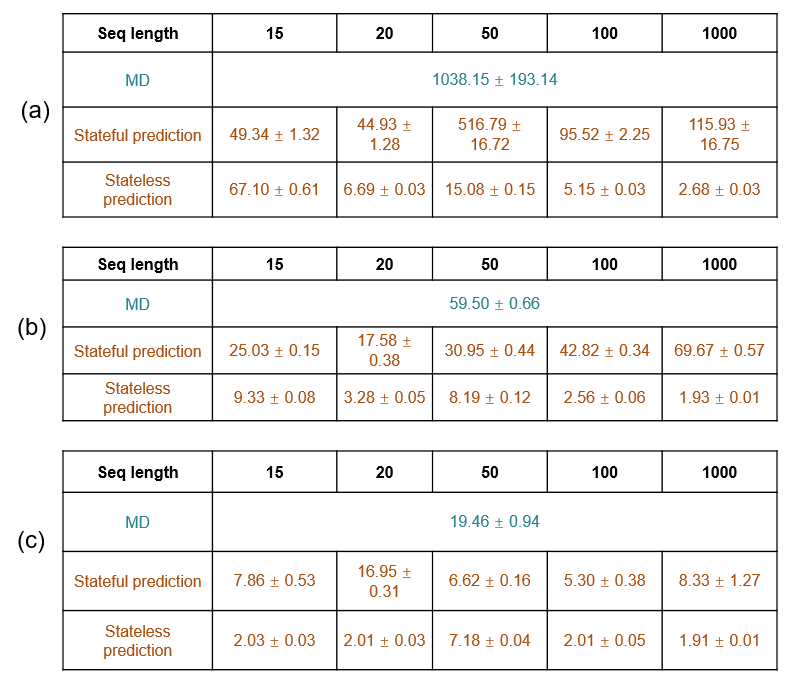}
\caption{(a)(b)(c) are 1st, 2nd, 3rd ITS on 0.1ps $alanine_{RMSD}$ respectively. Different sequence length does not help to improve. Stateful predictions outperforms stateless predictions.}
\label{stateful_all}
\end{table}

\begin{table}[hbp]   
\centering
\includegraphics[scale=0.5]{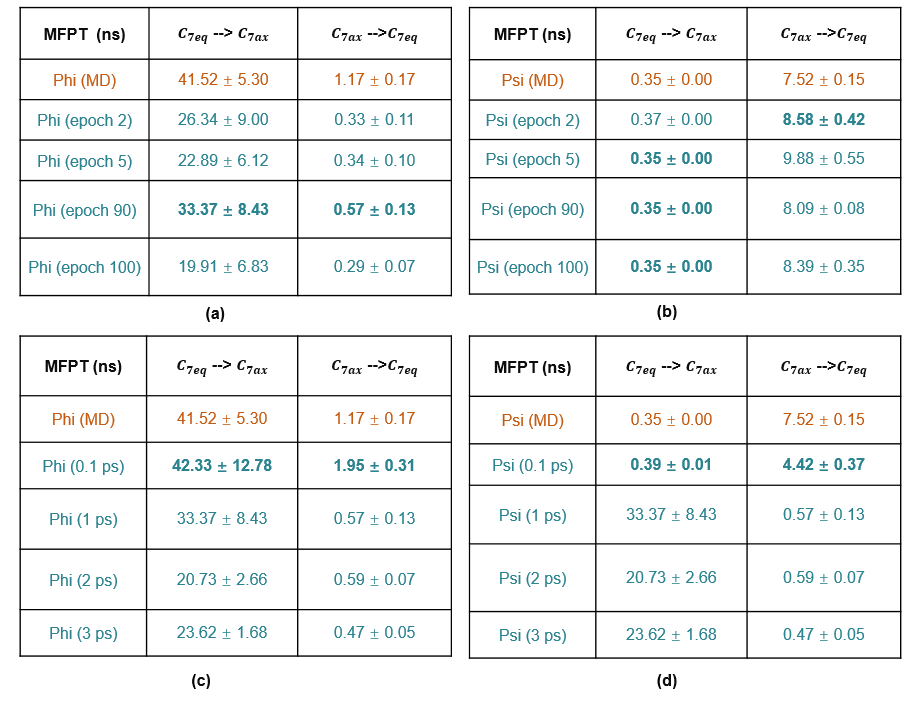}
\caption{MFPT for Transformer on $alanine_{\phi \psi}$ dataset. std gets calculated on the basis of 50 times bootstrap. (a)(b) are from 1ps saving interval with different epochs while (c)(d) are from epoch 90 with different saving intervals. Groundtruth is set from Molecular Dynamic simulation with saving interval=1ps. Different saving interval or epochs would lead to different MFPT results although their converged loss would be very much close as shown in Figure \ref{loss}. }
\label{mfpt}
\end{table}
 
\begin{table}[hbp] 
\centering
\includegraphics[scale=0.5]{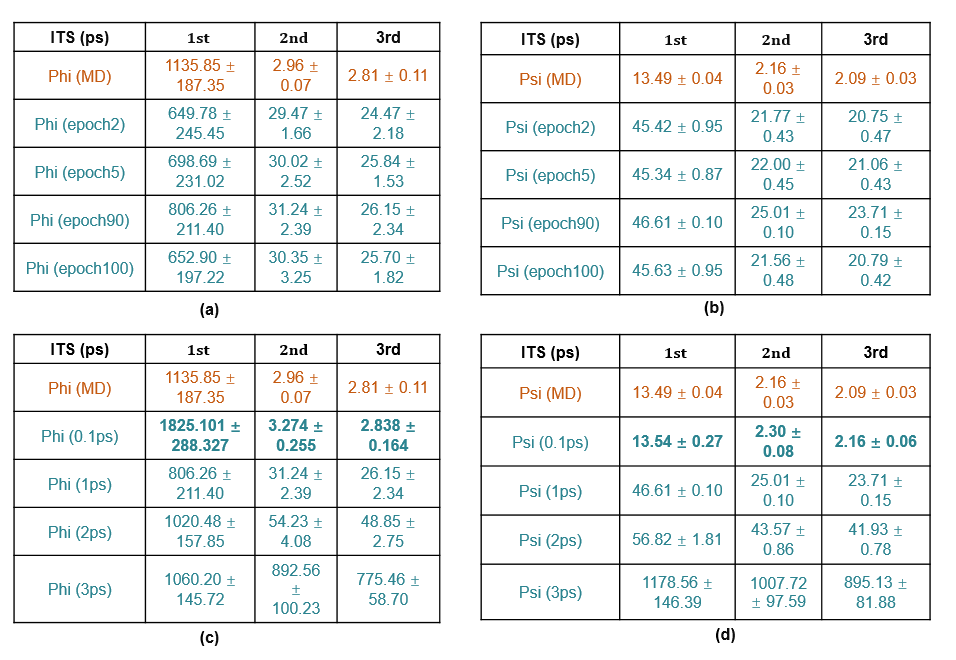}
\caption{ITS for Transformer on $alanine_{\phi \psi}$ dataset. std gets calculated on the basis of 50 times bootstrap. (a)(b) are from 1ps saving interval with different epochs while (c)(d) are from epoch 90 with different saving interval. Groundtruth is set from Molecular Dynamic simulation with saving interval=1ps. Different saving interval or epochs would lead to different ITS results although their converged loss would be very much close as shown in Figure \ref{loss}. }
\label{its}
\end{table}

\begin{table}[hbp] 
\centering
\includegraphics[scale=0.5]{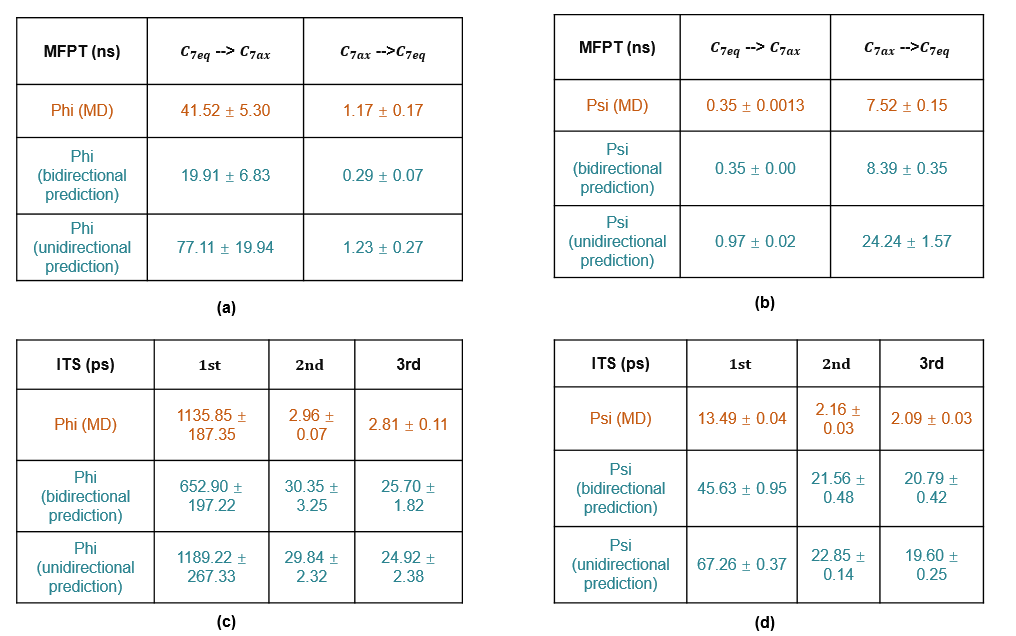}
\caption{(a)(b): MFPT on $\phi$ and $\psi$ respectively (c)(d): ITS on $\phi$ and $\psi$ respectively for unidirectional and bidirectional Transformer predictions on $alanine_{\phi \psi}$ with saving interval=0.1ps. Different but both poor performance.}
\label{uni bi all}
\end{table}

\end{document}